\documentclass[10pt,twocolumn,letterpaper]{article}

\usepackage{cvpr}              
\usepackage{textcomp}
\usepackage{stfloats}
\usepackage{url}
\usepackage{verbatim}
\usepackage{graphicx}
\usepackage{booktabs}
\usepackage{multirow}
\usepackage{epsfig}
\usepackage{graphicx}
\usepackage{amsmath}
\usepackage{amssymb}
\usepackage{bbding}
\usepackage{enumerate}
\usepackage{bbm}
\usepackage{dsfont}
\usepackage{amsmath,amssymb} 
\usepackage{commath}
\usepackage{enumitem}
\usepackage{fancyhdr} 

\definecolor{cvprblue}{rgb}{0.21,0.49,0.74}
\usepackage[pagebackref,breaklinks,colorlinks,allcolors=cvprblue]{hyperref}

\def\paperID{0008} 
\def\confName{CVPR}
\def\confYear{2026}

\title{A Paradigm Shift: Fully End-to-End Training for\\Temporal Sentence Grounding in Videos}

\author{
Allen He\textsuperscript{1} \quad
Qi Liu\textsuperscript{2} \quad
Kun Liu\textsuperscript{3}\thanks{Corresponding author.} \quad
Xinchen Liu\textsuperscript{3} \quad
Wu Liu\textsuperscript{4} \\
\textsuperscript{1}BASIS International School Park Lane Harbour \quad
\textsuperscript{2}UCAS \quad
\textsuperscript{3}JD Explore Academy \quad
\textsuperscript{4}USTC \\
{\tt\small allenhethis@outlook.com, Liukun167@jd.com} 
}

\begin{document}
\maketitle
\begin{abstract}

Temporal sentence grounding in videos (TSGV) aims to localize a temporal segment that semantically corresponds to a sentence query from an untrimmed video. 
Most current methods adopt pre-trained query-agnostic visual encoders for offline feature extraction, and the video backbones are frozen and not optimized for TSGV. 
This leads to a task discrepancy issue for the video backbone – trained for visual classification, but utilized for TSGV. 
To bridge this gap, we propose a fully end-to-end paradigm that jointly optimizes the video backbone and localization head. We first conduct an empirical study validating the effectiveness of end-to-end learning over frozen baselines across different model scales. 
Furthermore, we introduce a Sentence Conditioned Adapter (SCADA), which leverages sentence features to train a small portion of video backbone parameters adaptively. 
SCADA facilitates the deployment of deeper network backbones with reduced memory and significantly enhances visual representation by modulating feature maps through precise integration of linguistic embeddings.
Experiments on two benchmarks show that our method outperforms state-of-the-art approaches.
The code and models will be released.

\end{abstract}    
\section{Introduction}
\label{intro}
Joint modeling of videos and language, such as video captioning~\cite{xu2017learning,pan2017video} and video question answering~\cite{xu2017video,fan2019heterogeneous}, has gained significant attention in the multimedia community. 
In this work, we focus on Temporal Sentence Grounding in Videos (TSGV)~\cite{anne2017localizing,liu2018attentive}, which aims to localize activities in untrimmed videos through language queries~\cite{gao2017tall}. 
TSGV has broad applications in surveillance retrieval, movie search, and intelligent logistics.

\begin{figure}[t]
\centering
\includegraphics[width=\linewidth]{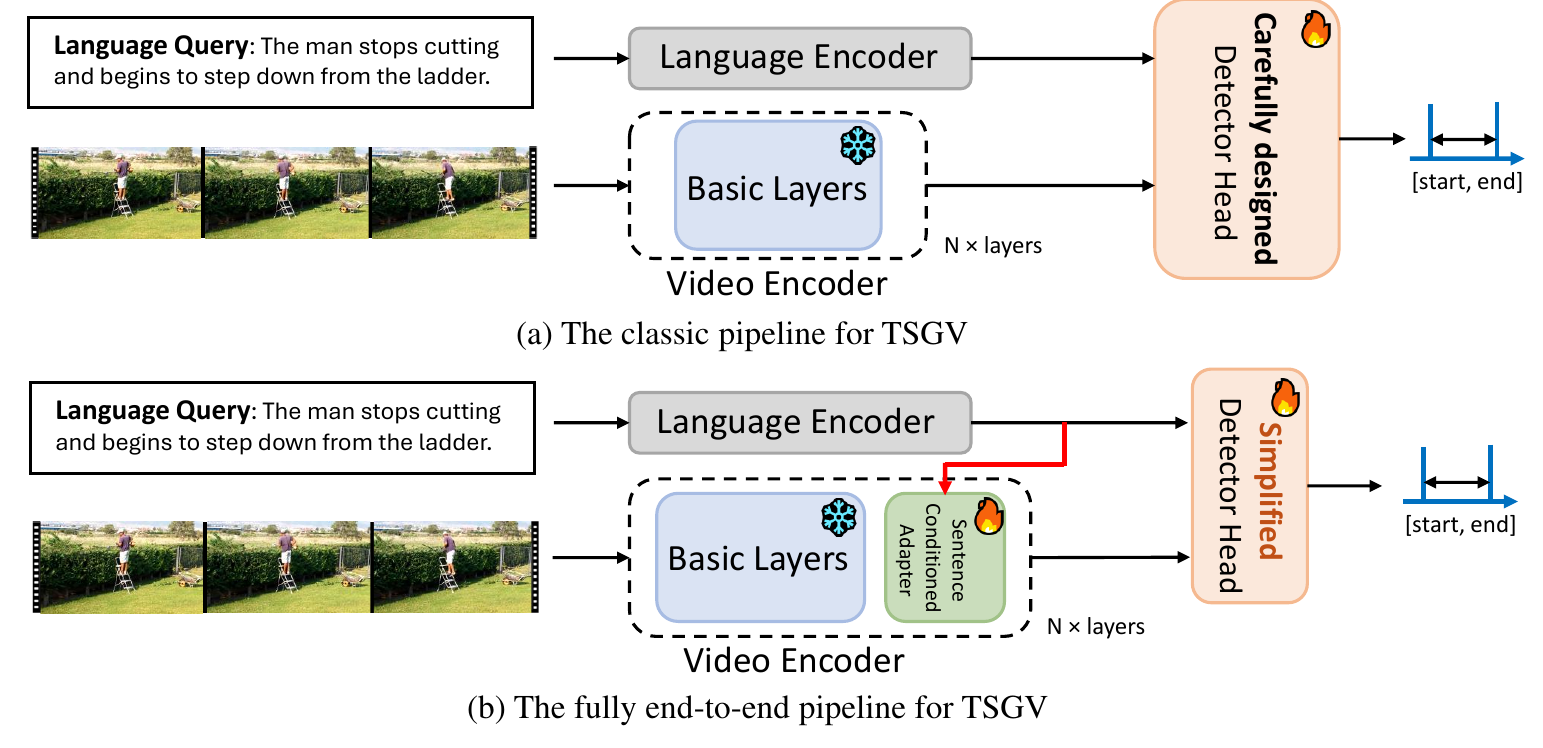}
\caption{An overview of the classic pipeline for TSGV (top) vs ours fully end-to-end pipeline (bottom).
}

\label{fig:01}
\end{figure}

Despite the promising results, the majority of current methods restrict the performance due to the sub-optimal video features and insufficient utilization of language. 
Most of the current TSGV pipeline mainly consist of the following stages, as shown in Figure ~\ref{fig:01}(a):
feature extraction stage to encode the videos and sentences into embedding,
cross-modal fusion stage to integrate the information from both modalities,
and moment localization stage to pinpoint the start and end time of the described activity within the video. 
First of all, the recent literature first pre-train the video encoder on a large video classification dataset then freezes it to extract features. 
After that, only the localization module upon the features is updated for the TSGV task. 
However, these backbones are typically trained on a limited range of objects or actions described at the phrase level. 
Consequently, pre-trained models are not likely to grasp the intricate semantics of natural language queries. 
Second, some state-of-the-art methods not only extract features from the two modalities independently but also fail to utilize sentence features during the localization stage. 
Although these methods integrate the language and video in the fusion stage, it is far from adequate to align the vision and sentences for TSGV due to the high requirement for the precise semantic alignment between the visual and natural language domains.

To address these limitations, as illustrated in Figure \ref{fig:01}(a), we propose a fully end-to-end framework that deeply integrates language throughout the entire pipeline, dynamically modulating both the visual backbone and the temporal localization head. 
Breaking away from the sub-optimal paradigm of frozen, independent feature extraction, we introduce a Sentence Conditioned Adapter (SCADA). 
As illustrated in Figure \ref{fig:01}(b), SCADA conditions the pre-trained visual backbone (e.g., ViT) on the sentence embedding, enabling the network to extract visual concepts tightly aligned with the query's intricate semantics.
To further enrich this visual learning and accelerate training, we reformulate the mini-batch construction to process multiple sentence queries per video simultaneously. 
Finally, rather than limiting language utilization to a standard cross-modal fusion stage, our method leverages the sentence embeddings to generate channel-wise attention weights, explicitly guiding the localization head to predict precise temporal boundaries. 
Extensive experiments demonstrate the superiority of our approach, achieving Rank1@IoU=0.5 scores of $48.1\%$ on Charades-STA and $30.5\%$ on ActivityNet-Captions.

To summarize, our contributions are as follows:
 \begin{itemize} 
    \item We introduce a novel, fully end-to-end framework for TSGV that uniquely leverages sentence embeddings to continuously guide the entire video grounding process.

    \item We propose a Sentence Conditioned Adapter (SCADA). It dynamically conditions the visual backbone on sentence embeddings to capture query-aligned visual concepts.

    \item Extensive experiments on two challenging benchmarks demonstrate that our method significantly outperforms existing state-of-the-art approaches. 
\end{itemize}
\section{Related Work}
\label{relate}
The previous related studies are mainly divided into two categories:
temporal activity localization, and temporal sentence grounding in videos.

\textbf{Temporal activity Localization.} 
Initially proposed by Gaidon et al.~\cite{gaidon2011actom}, this task detects the temporal boundaries of predefined activities. Existing methods primarily include computationally expensive two-stage approaches using sliding windows~\cite{shou2016temporal,zhao2017temporal}, and more direct one-stage single-shot methods~\cite{lin2017single,xu2017r,liu2022end}. Recently, end-to-end training mechanisms have further improved detection performance~\cite{liu2023etad,liu2023end,liu2022end,liu2022empirical}. Despite this progress, traditional localization is limited to a fixed set of word-level activities, struggling to capture the complex and diverse semantic concepts of real-world scenarios. This critical limitation has motivated the recent shift towards activity localization via natural language, which is the focus of our work.

\textbf{Temporal Sentence Grounding in Videos.} 
TSGV~\cite{anne2017localizing,gao2017tall} methods are primarily grouped into three categories: two-stage models, one-stage methods, and Video Large Language Models (Video LLMs).

Two-stage models~\cite{anne2017localizing,gao2017tall,liu2018attentive,ge2019mac,song2018val,jiang2019cross,zhang2019exploiting, lin2020moment, lu2024zero} first generate moment candidates via sliding windows and then match them with sentences. Despite enhancements like temporal regression~\cite{gao2017tall}, memory attention~\cite{liu2018attentive}, and verb-object mining~\cite{ge2019mac,jiang2019cross}, these pipeline-based methods remain computationally inefficient.

To improve efficiency, one-stage models~\cite{yang2024dynamic, chen2018temporally, 2DTAN, ning2021interaction, zhang2021multi, DBLP:conf/aaai/Wang0J20, yuan2019find, DBLP:conf/cvpr/ZengXHCTG20, DBLP:conf/aaai/XiaoCZJSYX21, DBLP:conf/aaai/LiGW21, DBLP:conf/cvpr/LiuQDZ00XX21, DBLP:conf/cvpr/ZhaoZZL21, DBLP:conf/emnlp/LiuQDZ21, DBLP:journals/tmm/YangWSL24, DBLP:journals/corr/abs-2205-05854, DBLP:conf/aaai/LiuQDCXZ22, plugmark, DBLP:conf/sigir/SunWGLZ22, DBLP:conf/emnlp/GaoSXZG21, DBLP:conf/emnlp/LiuQZ21, wang2022negative, DBLP:conf/emnlp/LuCTLX19, TMLGA, DBLP:conf/naacl/GhoshAPH19, DBLP:conf/cvpr/ZhouZLCH21, DBLP:conf/cvpr/WangZL0L21, liu2024sigformer, liu2024single, DBLP:conf/iscas/ZhengLYZLZ021, DBLP:conf/cvpr/Ma0WLLW24, DBLP:conf/mm/ChenLL0H022, DBLP:conf/icdsc/LiuMFZ14} directly predict temporal boundaries in a single pass. They utilize diverse mechanisms to achieve better alignment, including fine-grained word-frame interactions~\cite{chen2018temporally}, 2D Temporal Adjacent Networks (2D-TAN)~\cite{2DTAN}, biaffine architectures~\cite{DBLP:conf/cvpr/LiuQDZ00XX21}, joint embedding spaces~\cite{wang2022negative}, memory-guided learning~\cite{DBLP:conf/aaai/LiuQDCXZ22}, and coarse-to-fine exploration~\cite{yang2024dynamic, yuan2019find}.

Recently, Video LLMs have emerged for TSGV. Methods like TimeChat~\cite{TimeChat}, VTimeLLM~\cite{VTimeLLM}, HawkEye~\cite{HawkEye, InternVideo}, and others~\cite{LITA, Momentor, VTG-LLM} predict timestamps by leveraging time-sensitive instruction tuning, specialized tokens, or lightweight token compression.

While some recent work~\cite{zhang2023text} explores end-to-end optimization with a 2D visual encoder, its performance lags behind 3D backbone methods. In contrast, our framework optimizes a 3D visual backbone and detection head entirely end-to-end. By explicitly using sentence queries to modulate the 3D backbone, we achieve a deeper cross-modal fusion that surpasses current state-of-the-art approaches.

\textbf{Video Object Segmentation and Grounding.} 
Beyond purely temporal localization, grounding language in videos has also been widely explored at the pixel level~\cite{ding2025multimodal}. 
General video object segmentation in complex environments has been significantly advanced by benchmarks like MOSE~\cite{ding2023mose} and MOSEv2~\cite{ding2025mosev2}. Furthermore, multi-modal video object segmentation, which requires identifying and tracking objects based on motion expressions, has been deeply investigated in MeViS~\cite{ding2023mevis} and MeViSv2~\cite{ding2025mevis}.
\section{The Proposed Framework}
\label{thepr}

In this section, we first introduce the basic formation of TSGV.
We then describe the overall framework of our method.
Finally, we introduce the main parts of our method:
feature extraction stage, the sentence conditioned adapter, video-centic learning, and the detection head.

\subsection{Problem Definition}
Given an untrimmed video $V$ and a natural language query $S$, TSGV aims to localize the target moment $M$ that best aligns with $S$. Formally, the video is represented as a set of moment candidates $V = \{V_i\}_{i=1}^{l_v}$, where $l_v$ is the number of proposals. The query is defined as a word sequence $S = \{s_j\}_{j=1}^{l_s}$, where $l_s$ denotes the sentence length. During training, utilizing the ground-truth temporal boundaries, the network is optimized to predict the Intersection over Union (IoU) between each candidate and the target moment. At inference, the temporal grounding process is formulated as:$M = \mathop{\arg\max}_{V_i \in V} \sigma(V_i, S), \label{equ:01}$, where $\sigma(\cdot)$ denotes the trained network estimating the IoU score of candidate $V_i$ conditioned on query $S$.

\begin{figure*}[t]
  \centering 
  \includegraphics[width=0.98\linewidth]{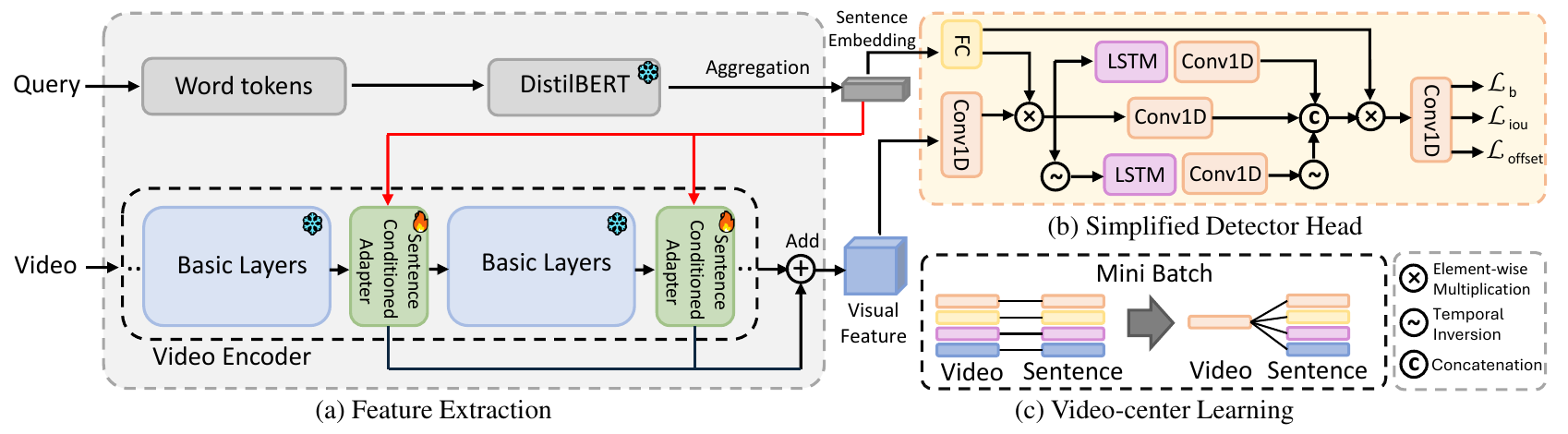}
\caption{
The proposed overall framework for fully end-to-end training for TSGV.
(a) In the Sentence Conditioned Adapter, sentence embeddings guide backbone fine-tuning to enhance visual feature extraction. 
Efficiency is improved via a (b) simplified detector head and (c) video-center learning.
}
\label{fig:02}
\end{figure*}

\subsection{Overall Framework}
Figure \ref{fig:02} illustrates our proposed end-to-end framework. During the feature extraction phase, sentence embeddings encoded by DistilBERT~\cite{DistilBERT} are utilized by the proposed SCADA module, which is interleaved within the visual backbone. This mechanism dynamically modulates the visual features conditioned on the text query, enabling highly effective visual fine-tuning with minimal computational overhead. For temporal localization, we adopt a streamlined detection head to maintain architectural efficiency. Furthermore, to accelerate the end-to-end training process and enrich visual feature optimization, we reformulate the mini-batch construction to simultaneously pair each video with multiple sentence queries.

\subsection{Feature Extraction Stage}

\textbf{Sentence Encoder.} We employ DistilBERT~\cite{DistilBERT} for lightweight language modeling. Given a query $S$, we extract a token-level feature sequence $\{f_i^S\}_{i=1}^{l_s+1} \in \mathbb{R}^{d^S}$ ($d^S=768$). The global sentence embedding is then derived by applying average pooling across all tokens.

\textbf{Video Encoder.} To comprehensively evaluate our framework, we explore three representative video architectures: C3D~\cite{tran2015learning}, I3D~\cite{carreira2017quo}, and Vision Transformers (ViT)~\cite{maev2}. While 3D convolutional networks (C3D and I3D) are standard baselines in TSGV, the potential of ViTs remains largely underexplored in this domain. Therefore, we integrate self-supervised pre-trained ViTs of varying capacities (e.g., ViT-S, ViT-B, and ViT-g) to extract robust spatial-temporal visual representations.

\begin{figure}[t]
  \centering
    \includegraphics[width=0.85\linewidth]{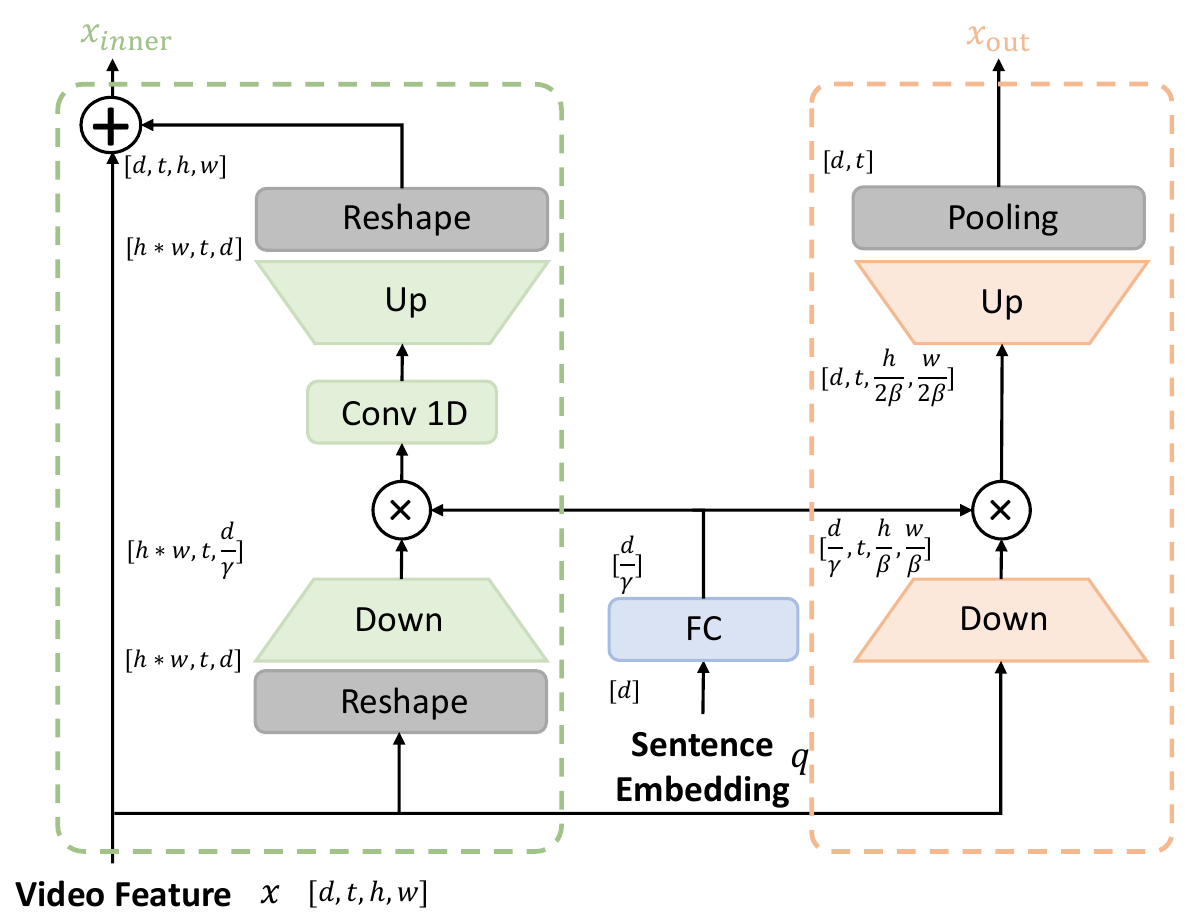}
\caption{The architecture of sentence conditioned adapter. 
It contains of inner and out branches that leverage sentences to guide the fine-tuning of the visual backbone network, ultimately enhancing the quality of the visual features.}
\label{fig:scada}
\end{figure}

\subsection{Sentence Conditioned Adapter}

Jointly fine-tuning a large visual backbone and a detector head causes substantial memory overhead and risks catastrophic forgetting of pre-trained knowledge. To address this, we introduce the Sentence Conditioned Adapter (SCADA). SCADA significantly reduces training memory while enhancing visual representations by dynamically modulating feature maps with linguistic embeddings.

As shown in Fig.~\ref{fig:scada}, given an intermediate visual feature $x \in \mathbb{R}^{d \times t \times h \times w}$ and a sentence query $q \in \mathbb{R}^d$, SCADA operates through two complementary pathways: an inner branch and an outer branch.

\textbf{Inner Branch.} This branch injects linguistic context into the visual stream and passes the output to the next backbone layer:
\begin{equation}
\begin{aligned}
   x' &= \sigma(\mathrm{FC}_{down}(\mathrm{Reshape}(x))), \\
   x'' &= \mathrm{DWConv1D}(\mathrm{Normalize}(x' \otimes \mathrm{FC}_{sentence}(q))), \\
   x_{inner} &= x + \mathrm{Reshape}(\sigma(\mathrm{FC}_{up}(x'')))
\end{aligned}
\label{equ:scada1}
\end{equation}
It first reduces the channel dimension of $x$ by a ratio $\gamma$. Next, it modulates the visual feature with $q$ via element-wise multiplication ($\otimes$) and applies a depth-wise 1D convolution to capture local temporal context. Finally, an FC layer restores the dimension, adding the residual $x$ to produce $x_{inner}$.

\textbf{Outer Branch.} This branch extracts query-guided, spatially compressed features that bypass the remaining backbone layers:
\begin{equation}
\begin{aligned}
   x'_{out} &= \sigma(\mathrm{Conv3D}_{down}(x)), \\
   x''_{out} &= \mathrm{Normalize}(x'_{out} \otimes \mathrm{FC}_{sentence}(q)), \\
   x_{out} &= \mathrm{Pooling}(\sigma(\mathrm{Conv3D}_{up}(x''_{out})))
\end{aligned}
\label{equ:scada2}
\end{equation}
It employs 3D convolutions to reduce the channel dimension and compress the spatial resolution by a ratio $\beta$. Following language modulation, a second 3D convolution restores the channels while further condensing the spatial size. Spatial pooling then yields $x_{out}$.

\textbf{Final Aggregation.} Rather than discarding the outer branch outputs, we aggregate them with the final backbone output $x_b$ to form the ultimate query-modulated visual representation:
\begin{equation}
F = \mathrm{Normalize}\left(x_b + \sum_{i=1}^n x_{outer}^i\right),
\label{equ:scada_final}
\end{equation}
where $n$ is the number of the SCADA.

\subsection{Detector Head and Loss Function}

\textbf{Detector Head.} To minimize computational costs during end-to-end training, we adopt a streamlined detector head inspired by~\cite{liu2023etad}. As shown in Fig.~\ref{fig:02}(b), it utilizes a bidirectional LSTM architecture with intermediate residual connections to capture temporal dependencies while preventing information loss over long sequences. To deeply integrate the two modalities, we repeatedly fuse the visual features with the sentence embeddings via element-wise multiplication throughout the detection process.

\textbf{Loss Function.} Our framework is optimized jointly using a multi-task loss defined as:
$$ \mathcal{L} = \mathcal{L}_{b} + \mathcal{L}_{iou} + \mathcal{L}_{offset} $$
Here, the \textbf{boundary loss} $\mathcal{L}_{b}$ applies a positive-negative-balanced binary cross-entropy loss~\cite{BMN} to supervise the start and end probabilities. This effectively handles the typical class imbalance between sparse boundary frames and frequent background frames. The \textbf{IoU loss} $\mathcal{L}_{iou}$ evaluates the overlap between predicted proposals and the ground truth, combining a classification term (via balanced cross-entropy) and a regression term (via L2 loss). Finally, the \textbf{offset loss} $\mathcal{L}_{offset}$ utilizes a smoothed L1 loss to fine-tune the temporal boundaries by regressing the fine-grained coordinate offsets (start, end, center, and width).

\subsection{Video-Center Learning}

Standard TSGV methods randomly sample individual video-query pairs to construct a mini-batch. In an end-to-end framework, this naive sampling is highly inefficient: because a single video often corresponds to multiple queries in the dataset, the network is forced to repeatedly dynamically extract features for the same video across different batches, causing severe computational redundancy.

To accelerate training, we propose a video-center learning strategy. As shown on the right side of Fig.~\ref{fig:02}(c), instead of sampling independent pairs, we first uniformly sample a video $V$ and group all its associated sentence queries $S^V = \{S_i \mid \langle V, S_i \rangle\}$ into the same mini-batch. For a batch size of $B$, this video-level sampling is simply repeated $\frac{B}{|S^V|}$ times. 

This strategy ensures that the visual backbone extracts the features of a given video only once per batch, sharing them across multiple queries. Beyond drastically reducing training time, this approach enriches the visual optimization process by forcing the network to simultaneously align the same video with various linguistic contexts in a single iteration.
\section{Experiments}
\label{exper}
To evaluate our approach, we conduct experiments on two popular and challenging benchmarks:
Charades-STA \cite{gao2017tall} and ActivityNet Captions~\cite{ANet}.
In this section, we first describe the two datasets, evaluation metrics, and implementation details.
Then, we compare the performance of our method with the state-of-the-art models.
Subsequently, we investigate the substantial improvements garnered through end-to-end learning and examine the impact of various design decisions on the TSGV task. 
These factors encompass the selection of the video encoder, the resolution of the input video, and the number of the frames sampled.
Finally, ablation studies are introduced to further demonstrate the effectiveness of our method.

\subsection{Datasets}
\label{data}

We evaluate our proposed method on two widely used TSGV benchmarks: Charades-STA and ActivityNet Captions.



\textbf{Charades-STA}~\cite{gao2017tall, sigurdsson2016hollywood} focuses on daily indoor activities. It includes 12,408 training and 3,720 testing pairs. The dataset features short videos (30.6s on average), brief target moments (8.2s), and simple queries (6.2 words).

\textbf{ActivityNet Captions} is a large-scale, open-domain dataset. Following the standard split~\cite{zhang2021multi}, we use 37,417 pairs for training, 17,505 for validation, and 17,031 for testing. Compared to Charades-STA, it is much more challenging due to longer videos (117.6s on average), longer target moments (37.1s), and complex queries (14.4 words). These factors demand stronger cross-modal understanding.

\subsection{Evaluation Metrics}
Following previous work \cite{gao2017tall,2DTAN},
we adopt ``Rank n@ tIoU = m'' and ``mIoU'' metric to evaluate our method.
``Rank n@ tIoU = m'' is defined as the percentage of sentence queries having at least one correct moment retrieval in the top $n$ retrieved clips.
A retrieved clip is correct when its tIoU (temporal Intersection Over Union)
with the target clip is larger than $m$.
Specifically, we use n $\in$ \{1, 5\} with m $\in$ \{0.5, 0.7\} for both datasets.
Meanwhile, ``mIoU'' means the average IoU for all the sentence queries.

\subsection{Implementation Details}
We utilize standard video feature extractors, such as C3D pre-trained on Sports1M and I3D pre-trained on Kinetics.
The entire frame sequence is fed into the video backbone, followed by spatial pooling to extract the visual features.
The dimension of detector head's input is $512$.
We fix the parameters of the visual backbone network and set a learning rate of $1 \times 10^{-3}$ for optimizing SCADA and the detector head.
The networks are optimized by AdamW optimizer
and the whole system is implemented via the PyTorch toolkit \cite{paszke2017automatic}.
All experiments are conducted on a single server with the NVIDIA A800 cards.

\begin{table*}[t]
\small
\centering
\setlength{\tabcolsep}{1mm}
\providecommand{\tabincell}[2]{\begin{tabular}{@{}#1@{}}#2\end{tabular}}

\caption{Comparison of state-of-the-art methods on Charades-STA and ActivityNet Captions dataset.}
\label{tab:chara_sota}
\vspace{1mm}

\begin{tabular}{c|c|ccccc|ccccc}
  \toprule
    \multirow{2}*{Backbone} & \multirow{2}*{Methods}   & \multicolumn{5}{c|}{Charades-STA}   & \multicolumn{5}{c}{ActivityNet-Captions}  \\ 
    & &\tabincell{c}{Rank1@\\IoU0.5}  &\tabincell{c}{Rank1@\\IoU0.7}    &\tabincell{c}{Rank5@\\IoU0.5}  &\tabincell{c}{Rank5@\\IoU0.7}         &\tabincell{c}{mIoU} &\tabincell{c}{Rank1@\\IoU0.5}  &\tabincell{c}{Rank1@\\IoU0.7} &\tabincell{c}{Rank5@\\IoU0.5}  &\tabincell{c}{Rank5@\\IoU0.7} &\tabincell{c}{mIoU}\\
  \midrule
      \multirow{5}*{LLM-based} & Momentor~\cite{Momentor}	& 26.60 & 11.60 &- &- & 28.50 &23.00	&12.40 &-&-&29.30  \\
       &HawkEye~\cite{HawkEye}	&31.40	&14.50	&-&-&33.70 &29.30 & 10.70 &-&-& 32.70\\
       & VideoExpert~\cite{videoexpert} & 40.30 & 20.90 &-&-& 41.10 &- &- &- \\
       & TRACE~\cite{trace} & 40.30 &19.40&-&- &- & 37.70 & 24.00 &-&-& 39.00 \\
       & D2VLM~\cite{zeng2025factorized} &50.30 & 26.00 &-&- &- &- &- &-&-&- \\
    \hline
       \multirow{6}*{C3D} 
       
       &MS-2D-TAN~\cite{zhang2021multi} 	&41.10	&23.25	&81.53 &48.55 &- & 46.16 & 29.21 &78.80 &60.85& -\\
       &APGN~\cite{DBLP:conf/emnlp/LiuQDZ21}	&48.20	&29.37	&89.05 &58.49&-&- &- &-&-&-\\
       &CPN~\cite{DBLP:conf/cvpr/ZhaoZZL21}	&46.08	&26.05	&-	&-	&43.90 &45.10	&28.10	&-	&-	&45.70\\
       &MMN~\cite{wang2022negative}	&- &-&-&- &- &48.59	&29.26	&{79.50} &{64.76}&-\\
       &G2L~\cite{G2L} & - & - &-&-& - & 51.68 & 33.35&81.32 &\textbf{67.60} & - \\
       &\textbf{Ours}	&\textbf{50.44}	&\textbf{31.19}	&\textbf{89.27} &\textbf{61.32} &\textbf{45.49} & \textbf{52.69} & \textbf{34.73}&\textbf{82.60} &{66.67} & \textbf{46.75}\\
  \hline
      \multirow{5}*{I3D} 
      &VSLNet~\cite{DBLP:conf/acl/ZhangSJZ20}	&47.31	&30.19	&-	&-	&45.15 &43.22	&26.16	&-	&-	&43.19\\
       &MS-2D-TAN~\cite{zhang2021multi} 	&56.64	&36.21	&89.14 &61.13 &- &45.50	&28.28&79.36 &61.70& -\\
       &MGPN~\cite{DBLP:conf/sigir/SunWGLZ22}	&60.82	&41.16	&89.77 &64.73&- &- &-&-&- &-\\
       &DPQF~\cite{DBLP:journals/tmm/YangWSL24}	&62.20	&42.52	&-&-&55.03 &- &-&-&- &-\\
       &\textbf{Ours}	&\textbf{64.59}	&\textbf{43.65}	&\textbf{92.69} &\textbf{72.31}&\textbf{55.94} &\textbf{48.49} &\textbf{31.36} &\textbf{81.08} &\textbf{66.33}& \textbf{45.69}\\
  \bottomrule
\end{tabular}
\end{table*}

\subsection{Comparison with State-of-the-art Methods}

We compare our approach with recent state-of-the-art methods on Charades-STA and ActivityNet Captions. As shown in Table~\ref{tab:chara_sota}, the baselines are grouped into Video LLM-based methods and traditional models utilizing C3D or I3D backbones. Our end-to-end framework consistently achieves the best performance across the majority of metrics on both datasets.

First, compared to traditional methods (e.g., MS-2D-TAN~\cite{zhang2021multi}, MMN~\cite{wang2022negative}, DPQF~\cite{DBLP:journals/tmm/YangWSL24}), our method shows significant improvements. Most existing models extract visual and language features independently and only optimize the detection head or the late-fusion mechanism. In contrast, our approach jointly optimizes the visual backbone and the detection head. By using SCADA to modulate visual features with language embeddings early on, we achieve much tighter cross-modal alignment. For instance, on Charades-STA with the I3D backbone, we achieve 64.59\% and 43.65\% in Rank1@IoU0.5 and Rank1@IoU0.7, outperforming the strong baseline DPQF.

Second, our method demonstrates clear superiority over recent Video LLM-based methods (e.g., Momentor~\cite{Momentor}, HawkEye~\cite{HawkEye}, VideoExpert~\cite{videoexpert}). While LLMs excel at coarse-grained video understanding, they often struggle to predict precise temporal boundaries. By coupling early language guidance with a dedicated temporal detection head, our C3D-based model surpasses the best LLM method, VideoExpert, by 10.14\% in Rank1@IoU0.5 on Charades-STA.

Finally, we observe consistent success on ActivityNet Captions, which poses a greater challenge due to its longer videos and complex activities. Even in this demanding setting, our method using the C3D backbone reaches 52.69\% in Rank1@IoU0.5, surpassing the recent G2L~\cite{G2L} (51.68\%). This confirms that integrating language cues to dynamically guide visual feature extraction is highly effective for localizing complex activities in long, untrimmed videos.

\renewcommand\arraystretch{1.2}
\begin{table*}[t]\footnotesize
\begin{center}
\newcommand{\tabincell}[2]{\begin{tabular}{@{}#1@{}}#2\end{tabular}}
\centering \caption{Comparison of E2E learning and non-E2E learning of different visual backbones on Charades-STA and ActivityNet Captions.}
\label{tab:e2e}
\vspace{1mm}
\begin{tabular}{c|c|c|ccc|c|ccc|c}
  \toprule
       \multirow{2}*{Backbone} & \multirow{2}*{Param.} & \multirow{2}*{E2E}  &\tabincell{c}{Rank1@\\IoU0.7} &\tabincell{c}{Rank5@\\IoU0.7} &\tabincell{c}{mIoU}  &\tabincell{c}{Average\\Gain}   &\tabincell{c}{Rank1@\\IoU0.7}          &\tabincell{c}{Rank5@\\IoU0.7} &\tabincell{c}{mIoU}  &\tabincell{c}{Average\\Gain}\\
  \cline{4-11}
  & & &\multicolumn{4}{c|}{\textbf{Charades-STA}} &\multicolumn{4}{c}{\textbf{ActivityNet Captions}}\\
  \midrule
    \multirow{2}*{C3D} &\multirow{2}*{27.66 M} & \XSolidBrush &21.99	&32.98	&32.31 &-  	&25.32	&57.23	&41.99 &-\\
    &&\Checkmark	&31.19	&61.32	&45.49 &+16.91 	&32.73	&66.67	&46.75 &+7.20\\
    \midrule
    \multirow{2}*{I3D} &\multirow{2}*{46.16 M} &\XSolidBrush &30.03	&48.84	&43.80 &- 	&26.58	&58.74	&43.18 &-\\
    &&\Checkmark		&42.65		&72.31	&55.94 &+16.08 	&31.36	&66.33	&45.69 & +4.96\\
    \midrule
     \multirow{2}*{ViT-S} &\multirow{2}*{21.88 M} &\XSolidBrush  &32.50	&50.86	&47.30 &- 	&28.90	&59.66	&44.56 & - \\
     &&\Checkmark	&45.48	&72.28	&56.19 &+14.43 &32.65	&66.71	&46.93 & +4.39\\
     \midrule
     \multirow{2}*{ViT-B} & \multirow{2}*{86.23 M} &\XSolidBrush  &39.65	&60.02	&49.57 &- &29.66	&61.31	&44.71 & -\\
     &&\Checkmark	&46.32	&73.21	&56.29& +8.86 &32.91	&67.58	&47.03 & +3.95\\
     \midrule
     \multirow{2}*{ViT-g} & \multirow{2}*{774.60 M} &\XSolidBrush &38.74	&63.85	&51.60 &- &29.21	&62.11	&45.05 & -\\
     &&\Checkmark	&48.98	&72.19	&57.78& +8.25 &33.64	&69.36	&48.27 & +4.97\\

  \bottomrule
\end{tabular}
\end{center}
\end{table*}

\subsection{The Effect of End-to-end Learning}
\subsubsection{E2E vs. Head-only}

In this section, we investigate the impact of end-to-end (E2E) training by comparing it against a ``Head-only'' baseline, which relies on offline features extracted from fixed backbones. We evaluate established architectures (C3D~\cite{tran2015learning}, I3D~\cite{carreira2017quo}) and further explore Vision Transformers (ViT-S~\cite{vit}, ViT-B~\cite{vit}, and ViT-g~\cite{vit-g}) pre-trained with VideoMAE V2~\cite{maev2}.

As shown in Table~\ref{tab:e2e}, E2E training consistently yields substantial improvements across all backbones and datasets. On Charades-STA, E2E training boosts the average performance of C3D, I3D, and ViT-S by 16.91\%, 16.08\%, and 14.43\%, respectively. Even the highly capable ViT-B and ViT-g models gain over 8\% with E2E optimization. While ActivityNet Captions is more challenging, E2E training still provides significant and consistent average gains ranging from 3.95\% to 7.20\% across all evaluated backbones. These results indicate that skipping E2E training causes a severe performance drop due to the divergence between standard action recognition pre-training and the specific requirements of TSGV.

Crucially, when using frozen offline features, our streamlined detection head performs poorly compared to existing methods. However, under the E2E paradigm, it achieves state-of-the-art results. This demonstrates that complex detection heads are unnecessary if the visual backbone is jointly optimized. By directly bridging the gap between pre-training tasks and TSGV, E2E learning provides highly task-specific visual features, allowing a simple and efficient detection head to deliver superior performance.

\subsubsection{The Effect of Training Augmentation}
We evaluate the impact of data augmentation within our end-to-end framework. Unlike offline feature extraction, end-to-end training allows for the seamless integration of both image and text augmentations during the learning process to prevent overfitting and improve generalization.

\renewcommand\arraystretch{1.2}
\begin{table}[t]\footnotesize
\setlength{\tabcolsep}{2.5pt}
\begin{center}
\newcommand{\tabincell}[2]{\begin{tabular}{@{}#1@{}}#2\end{tabular}}
\centering \caption{The effect of image augmentations.}
\label{tab:iamge_au}
\vspace{1mm}
\begin{tabular}{cccc|ccc}
  \toprule
       Cropping&  \tabincell{c}{Horizontal\\Flipping} & Rotation &  Distortion    &\tabincell{c}{Rank1@\\IoU0.5}  &\tabincell{c}{Rank1@\\IoU0.7}    &\tabincell{c}{mIoU}\\
  \midrule
    &&&& 45.13	&26.69		&41.42\\
    \Checkmark &&& &47.90	&27.85		&41.83\\
    \Checkmark & \Checkmark &&  &47.75	&28.01		&43.04\\
    \Checkmark & \Checkmark & \Checkmark & &48.65	&29.41		&43.63\\
    \Checkmark & \Checkmark & \Checkmark & \Checkmark &\textbf{50.44}	&\textbf{31.19}		&\textbf{45.49}\\
  
  \bottomrule
\end{tabular}
\end{center}
\end{table}

\renewcommand\arraystretch{1.2}
\begin{table}[t]\footnotesize
\setlength{\tabcolsep}{3.5pt}
\begin{center}
\newcommand{\tabincell}[2]{\begin{tabular}{@{}#1@{}}#2\end{tabular}}
\centering \caption{The effect of text augmentations.}
\label{tab:text_au}
\vspace{1mm}
\begin{tabular}{l|ccccc}
  \toprule
       Augmentation      &\tabincell{c}{Rank1@\\IoU0.5}  &\tabincell{c}{Rank1@\\IoU0.7}          &\tabincell{c}{Rank5@\\IoU0.5}  &\tabincell{c}{Rank5@\\IoU0.7} &\tabincell{c}{mIoU}\\
  \midrule
    Baseline & 45.13	&26.69	&83.82	&\textbf{57.63}	&41.42\\
    Swapping &45.43	&27.29	&\textbf{84.03}	&56.99	&41.03\\
    Insertion  &45.75	&26.77	&83.59	&57.04	&41.34\\
    Replacement &\textbf{46.51}	&\textbf{27.85}	&81.24	&57.26	&\textbf{42.45}\\
   
  \bottomrule
\end{tabular}
\end{center}
\end{table}

\textbf{Image Augmentations.}
We progressively apply random cropping, horizontal flipping, rotation, and photometric distortion (altering brightness, contrast, saturation, and hue). Using the C3D backbone on Charades-STA, the results in Table~\ref{tab:iamge_au} demonstrate that these visual enhancements significantly boost performance. While adding horizontal flipping to cropping causes a negligible dip in Rank1@IoU=0.5, combining all four augmentations yields the best overall results, increasing Rank1@IoU=0.5 from 45.13\% to 50.44\%. This confirms that image augmentation is highly effective for expanding small training datasets. Notably, even our baseline model without any augmentation outperforms traditional offline methods, further validating the inherent advantage of jointly optimizing the visual backbone and the localization head.

\textbf{Text Augmentations.} 
We also explore three text augmentation strategies: random word swapping, insertion, and synonym replacement. As shown in Table~\ref{tab:text_au}, text augmentation is notably less effective than image augmentation. Swapping and insertion fail to improve performance, likely because they disrupt natural sentence structure and hinder the model's language parsing. Synonym replacement yields modest gains (e.g., improving Rank1@IoU=0.5 by 1.38\% and Rank1@IoU=0.7 by 1.16\%) but causes a notable 2.58\% drop in Rank5@IoU=0.5. This suggests that while synonym replacement introduces useful linguistic diversity, text augmentations generally struggle to provide consistent improvements across all ranking metrics in this task.

\subsection{Evaluation of Design Choices} 
\subsubsection{Study on the Video Encoder}
\label{video_encoder}

As shown in Table~\ref{tab:e2e}, we analyze the impact of different visual backbones and observe two key findings.
(1) Scaling up the model size consistently improves performance.
Transitioning from C3D to the larger I3D backbone yields substantial gains. Similarly, upgrading the Transformer backbones from ViT-S to ViT-B and ViT-g further boosts accuracy. While the relative improvements for larger ViTs are slightly smaller under end-to-end training due to performance saturation, the overall upward trend remains clear.
(2) Self-supervised Vision Transformers outperform 3D convolutional networks.Despite having fewer parameters than both C3D and I3D, ViT-S achieves significantly better results. This highlights the strong representational power of Transformer architectures and the distinct advantages of self-supervised video pre-training for TSGV tasks.

\begin{figure}[t]
  \centering
  \includegraphics[width=\linewidth]{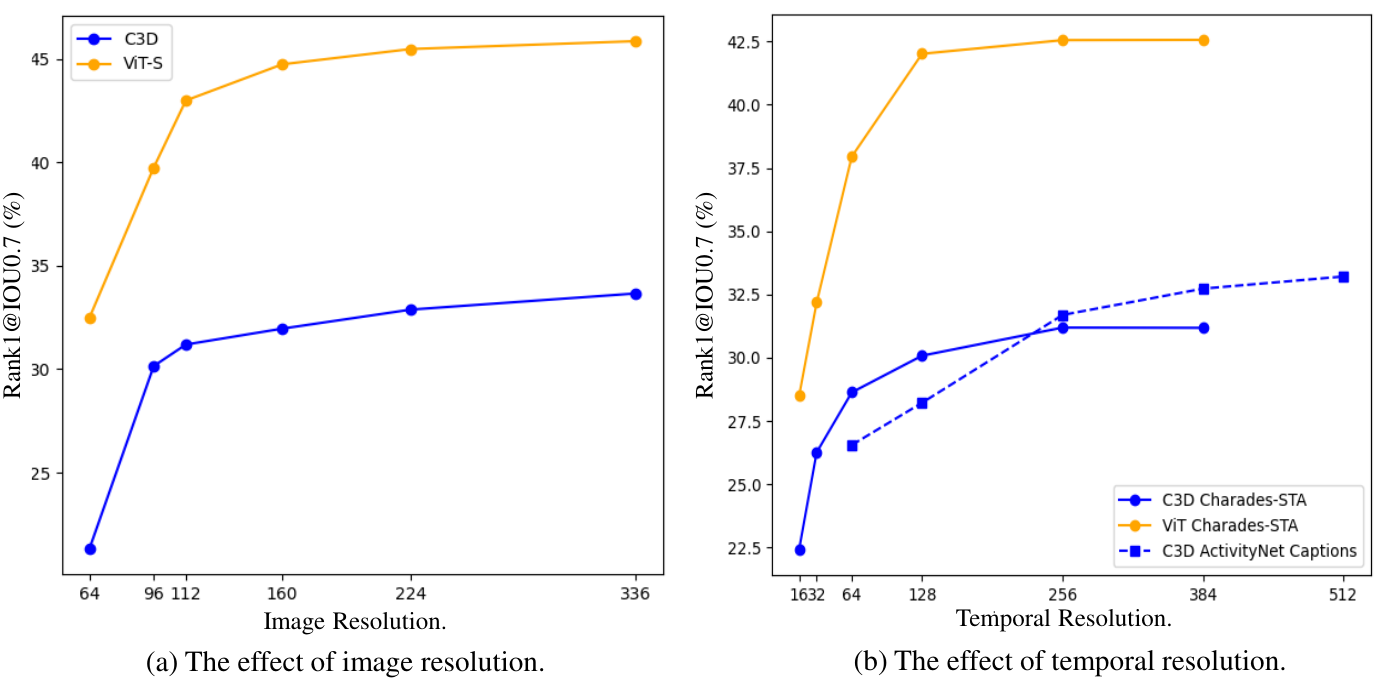}
\caption{The effect of image and temporal resolution.}
\label{fig:size}
\end{figure}

\renewcommand\arraystretch{1.2}
\begin{table}[t]\footnotesize
\begin{center}
\newcommand{\tabincell}[2]{\begin{tabular}{@{}#1@{}}#2\end{tabular}}
\centering \caption{Comparison of end-to-end learning and non-end-to-end learning using different video resolutions.}
\label{tab:96re}
\vspace{1mm}
\begin{tabular}{c|c|c|cc}
  \toprule
       Backbone  &E2E  & Video Res.   &\tabincell{c}{Rank1@\\IoU0.7}  &\tabincell{c}{Rank5@\\IoU0.7}\\
  \midrule
  \multicolumn{5}{c}{\textbf{Charades-STA}} \\
  \midrule
    \multirow{2}*{C3D} & \XSolidBrush & 112 &\textbf{23.25}	&48.55\\
    & \Checkmark & 64 &21.96	&\textbf{53.14}\\
    \midrule
    \multirow{2}*{I3D} &\XSolidBrush & 224 &36.21	&61.13\\
    & \Checkmark & 96  & \textbf{37.05} &	\textbf{63.74}\\
    \midrule
    \multicolumn{5}{c}{\textbf{ActivityNet Captions}} \\
    \midrule
    \multirow{2}*{C3D}  &\XSolidBrush &112 &29.21	&\textbf{60.85}\\
    & \Checkmark & 96 &\textbf{30.48}	 &60.19\\
    \midrule
    \multirow{2}*{I3D}  &\XSolidBrush & 224 &28.28	&61.70\\
    & \Checkmark &96 &\textbf{28.83}	&\textbf{63.49}\\
   
  \bottomrule
\end{tabular}
\end{center}
\end{table}

\subsubsection{Study on Image and Temporal Resolution}

We investigate the impact of spatial and temporal resolutions on TSGV performance, with results shown in Figure~\ref{fig:size}.

\textbf{Image Resolution.} First, increasing spatial resolution improves performance by capturing finer details, though gains diminish beyond a resolution of 112. Second, encoder architecture is more critical than input size: a ViT-S model at a low resolution of 64 matches the performance of a C3D model at 224. Finally, Table~\ref{tab:96re} demonstrates that end-to-end (E2E) training allows for drastic resolution reductions without sacrificing accuracy. For example, an E2E-trained I3D at a resolution of 96 matches or outperforms a frozen I3D at 224 across both datasets. On Charades-STA, the C3D resolution can even be dropped to 64.

\textbf{Temporal Resolution.} Increasing the number of sampled frames provides crucial temporal context, but excessive frames merely increase computational costs with diminishing returns. As shown in Figure~\ref{fig:size}(b), uniformly sampling 128 frames per video is optimal for Charades-STA. However, because ActivityNet Captions features significantly longer videos, sampling 384 frames is necessary to maintain sufficient temporal context for precise localization.

\subsection{Ablation Studies}
In this subsection, we perform detailed ablation studies on the Charades-STA and ActivityNet Captions to demonstrate the effects of the proposed framework components.

\renewcommand\arraystretch{1.2}
\begin{table}[t]\footnotesize
\setlength{\tabcolsep}{3.5pt}
\begin{center}
\newcommand{\tabincell}[2]{\begin{tabular}{@{}#1@{}}#2\end{tabular}}
\centering \caption{The effect of SCADA.
Param. is the number of tunable parameters in the backbone. 
``Full'' fine-tunes all backbone parameters and ``Partial'' updates only the final layers.
* means out of memory on A800-80GB.
}
\label{tab:scada}
\vspace{1mm}
\begin{tabular}{c|c|c|ccc}
  \toprule
       Backbone & Method  &Param.   &\tabincell{c}{Rank1@\\IoU0.5}  &\tabincell{c}{Rank1@\\IoU0.7}  &\tabincell{c}{mIoU}\\
 \midrule
  \midrule
    \multirow{4}*{ViT-S} & Full &21.88 M &64.01	&45.06	&55.64\\
    & Partial &4.14 M &64.17	&44.46	&55.13\\
    & SCADA w/o text &5.28 M &{63.21}	&{43.31}	&54.06\\
    & SCADA &5.28 M &\textbf{65.03}	&\textbf{45.48}	&\textbf{55.94}\\

  \midrule
    \multirow{3}*{ViT-g} & Full$^*$ &774.60 M &-	&-	&-\\
    &SCADA w/o text &13.21 M &{64.93}  &{46.38}		&{55.13} \\
    & SCADA &13.21 M &\textbf{66.53}  &\textbf{48.98}		&\textbf{57.78} \\
  \bottomrule
\end{tabular}
\end{center}
\end{table}

\subsubsection{The Effect of SCADA}

We evaluate the effectiveness and efficiency of SCADA using the ViT-S and ViT-g backbones. As detailed in Table~\ref{tab:scada}, our ablation analysis reveals two major advantages of our proposed module:

\textbf{Parameter Efficiency and Scalability.} While ``Full'' fine-tuning achieves competitive results on smaller models like ViT-S (reaching 64.01\% on Rank1@IoU0.5), it becomes computationally prohibitive for massive models. For instance, attempting to fully fine-tune the 774.60 M parameters of ViT-g results in out-of-memory (OOM) errors on an A800 GPU. ``Partial'' fine-tuning mitigates this memory bottleneck by updating only the final layers (e.g., training just 4.14 M parameters for ViT-S), but it sacrifices localization performance because the shallow visual features remain unadapted to the TSGV task. In contrast, SCADA introduces only a lightweight set of trainable parameters (5.28 M for ViT-S and 13.21 M for ViT-g) yet consistently outperforms both Full and Partial fine-tuning. For ViT-S, SCADA achieves a top Rank1@IoU0.5 score of 65.03\%. This demonstrates that SCADA successfully balances memory efficiency with superior feature representation, enabling the practical deployment of giant vision models in TSGV.

\textbf{The Importance of Language Modulation.} To verify the necessity of text-conditioned visual extraction, we test a variant that removes the language input (``SCADA w/o text''), essentially treating it as a standard visual-only adapter. As shown in Table~\ref{tab:scada}, excluding text causes a noticeable drop in performance across all metrics for both backbones. For the ViT-g model, introducing language modulation via SCADA increases Rank1@IoU0.5 from 64.93\% to 66.53\% and mIoU from 55.13\% to 57.78\%. This confirms that explicitly modulating the visual backbone with the sentence query is critical. It ensures that the encoder dynamically extracts task-specific visual features strictly aligned with the target semantics early in the pipeline, which significantly improves the final grounding accuracy compared to relying on isolated visual features.

\begin{figure}[t]

\centering
\includegraphics[width=\linewidth]{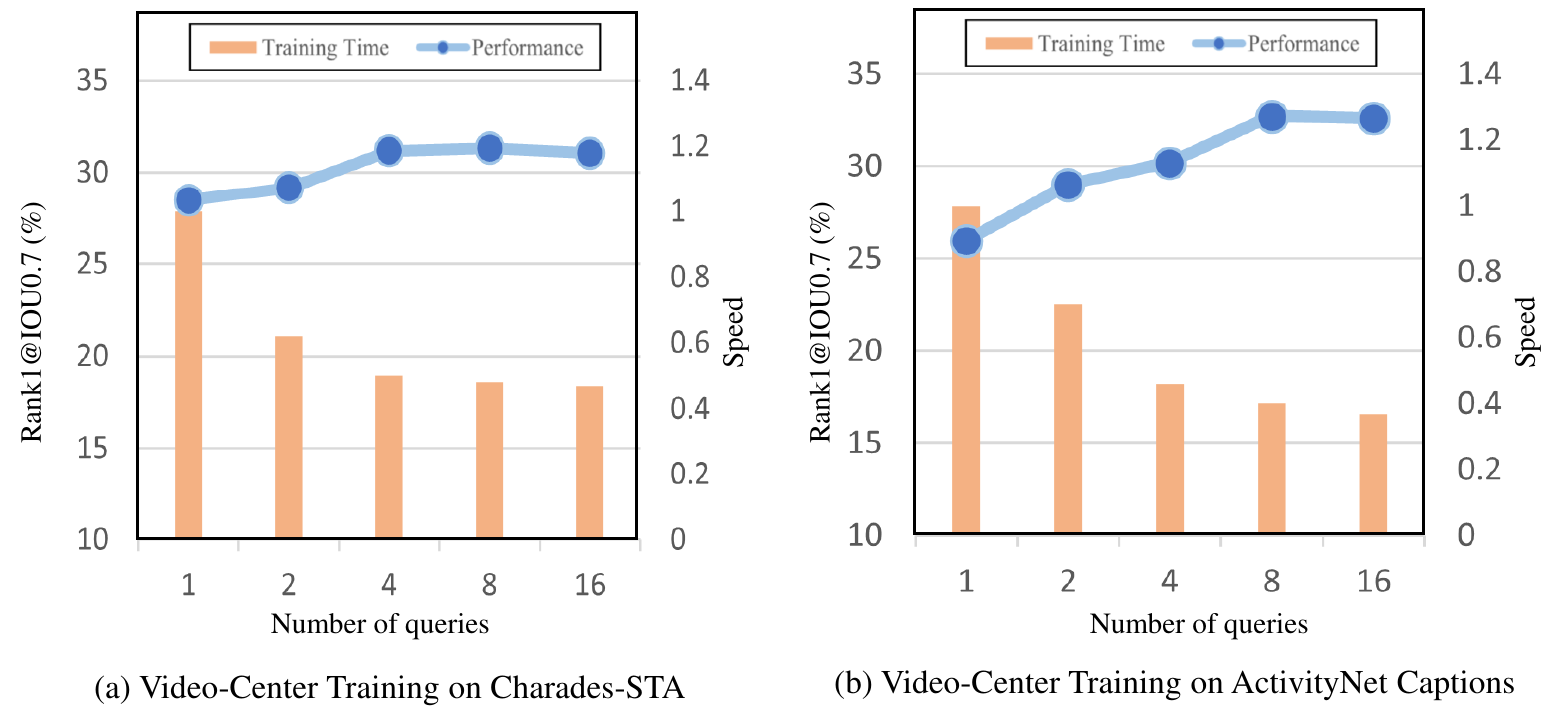}

\caption{Comparison of training time and test performance for the number of concurrent queries trained on a single video. 
The training time for a single query is taken as the baseline and normalized to 1. 
We report Rank1@IoU0.7 on both datasets using C3D as the backbone.}
\label{fig:sent}
\end{figure}

\subsubsection{The Effect of Video-Centric Training}

As shown in Figure~\ref{fig:sent}, we evaluate the impact of varying the number of queries per video on both training efficiency and grounding performance. We maintain a fixed total batch size of 32 (e.g., 4 videos paired with 8 queries each) and normalize the training time of a single video-query pair to 1 as our baseline.

On Charades-STA, increasing the queries per video from 1 to 4 significantly reduces training time and improves performance. However, because this dataset averages only 2.4 queries per video, further increasing the query count yields no additional gain. 
Conversely, ActivityNet Captions inherently contains more queries per video. Consequently, increasing the queries per video to 8 yields substantial improvements in Rank1@IoU0.7, while slashing the training time to approximately 40\% of the baseline. 
In conclusion, our video-centric training strategy drastically accelerates the end-to-end learning process while consistently enhancing overall localization performance.

\section{Conclusion}
\label{concl}
In this work, we address the task discrepancy in temporal sentence grounding by introducing the first fully end-to-end framework that jointly optimizes video backbones and localization heads. 
Our core innovation, the Sentence Conditioned Adapter (SCADA), enables language-guided visual feature modulation while maintaining efficiency through sparse parameter updates. Our approach not only achieves state-of-the-art performance but also reveals crucial insights: end-to-end training provides consistent gains across architectures, and strategic visual-linguistic fusion in early stages proves more effective than late-stage integration alone.
In the future, we would like to extend our method to other vision and language tasks,
such as video question answering, video captioning, etc.
\

{
    \small
    \bibliographystyle{ieeenat_fullname}
    \bibliography{main}
}

\end{document}